\documentclass[10pt,twocolumn,letterpaper]{article}

\usepackage{iccv}
\usepackage{times}
\usepackage{epsfig}
\usepackage{graphicx}
\usepackage{amsmath}
\usepackage{amssymb}
\usepackage{subfigure}
\usepackage{caption}
\usepackage{mathtools}
\usepackage{siunitx}
\usepackage{array}
\usepackage{multirow}
\usepackage{hhline}
\usepackage{array}
\newcommand{\argmin}{\operatornamewithlimits{argmin}}

\usepackage[pagebackref=true,breaklinks=true,letterpaper=true,colorlinks,bookmarks=false]{hyperref}
\iccvfinalcopy % *** Uncomment this line for the final submission

\def\iccvPaperID{1144} % *** Enter the ICCV Paper ID here

\ificcvfinal\fi
\begin{document}

%%%%%%%%% TITLE
\title{See the Difference: Direct Pre-Image Reconstruction and \\Pose Estimation by Differentiating HOG}

\author{Wei-Chen Chiu \qquad Mario Fritz\\
Max Planck Institute for Informatics, Saarbr\"{u}cken, Germany\\
{\tt\small \{walon,mfritz\}@mpi-inf.mpg.de}}

\maketitle

%%%%%%%%% ABSTRACT
\begin{abstract}
The Histogram of Oriented Gradient (HOG) descriptor has led to many advances in computer vision over the last decade and is still part of many state of the art approaches.
We realize that the associated feature computation is piecewise differentiable and therefore many pipelines which build on HOG can be made differentiable. This lends to advanced introspection as well as opportunities for end-to-end optimization. We present our implementation of $\nabla$HOG based on the auto-differentiation toolbox \textnormal{Chumpy \cite{Chumpy}} and show applications to pre-image visualization and pose estimation which extends the existing differentiable renderer \textnormal{OpenDR \cite{loper2014opendr}} pipeline. Both applications improve on the respective state-of-the-art HOG approaches.
\end{abstract}

%%%%%%%%% BODY TEXT
\section{Introduction}

\begin{figure}
\begin{center}
\includegraphics[width=1\columnwidth]{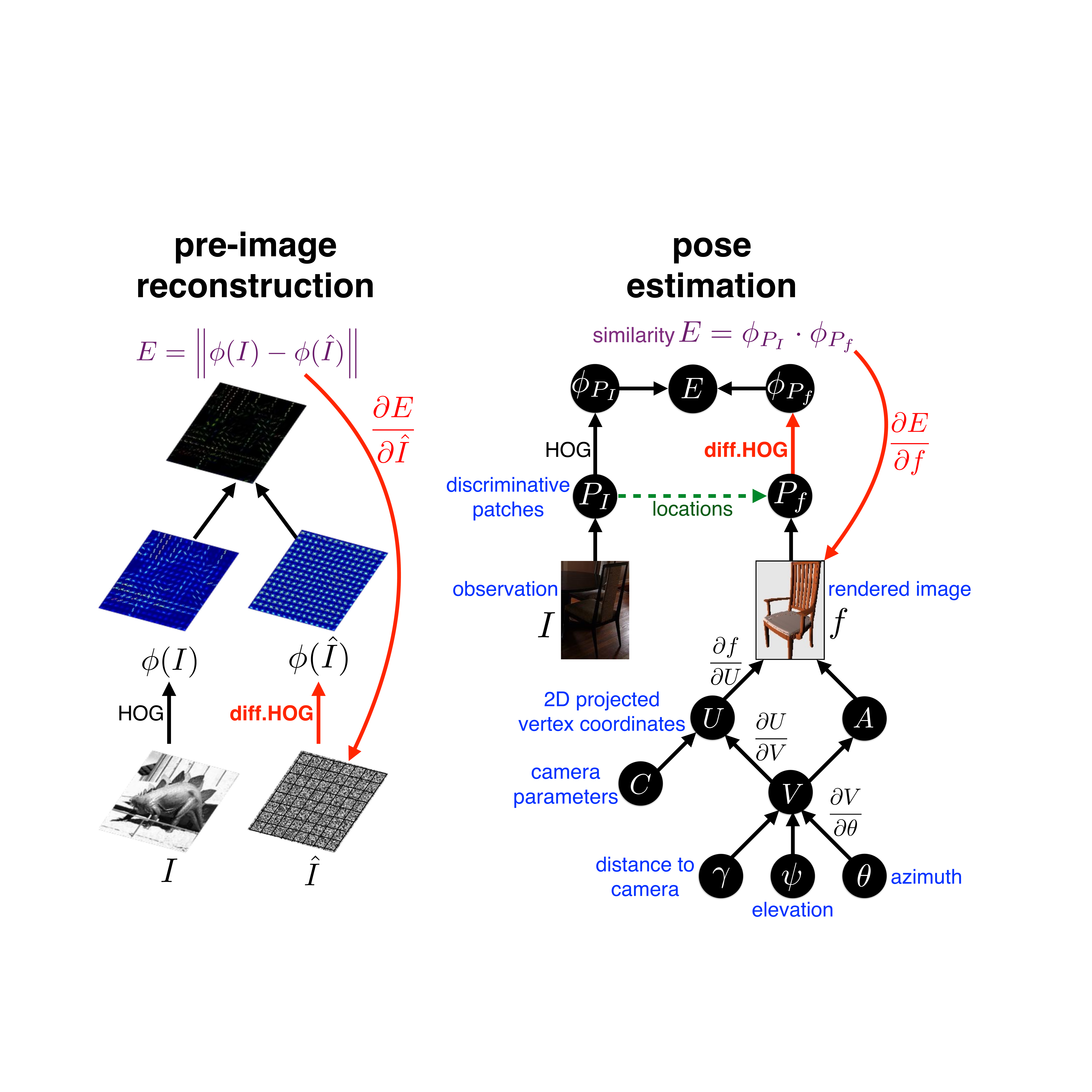}
\end{center}
\caption{We exploit the piecewise differentiability of the popular HOG descriptor for end-to-end optimization. The figure shows applications on the pre-image reconstruction given HOG features as well as the pose estimation task based on the same idea.}
\label{fig:teaser}
\end{figure}

Since the original presentation of the Histogram of Oriented Gradients (HOG) descriptor \cite{dalal2005histograms} it has seen many use cases beyond its initial target application to pedestrain detection. Most prominently it is a core building block of the widely used Deformable Part Model (DPM) object class detector \cite{felzenszwalb2010object} and exemplar models \cite{malisiewicz-iccv11} which both on their own have seen many follow-up approaches. Most recently, HOG-based approaches have repeatedly shown good generalization performance to rendered \cite{Aubry14} and artistic images \cite{Aubry13}, while such type of generalizations are non-trivial to achieve in recently very successful deep learning models in vision \cite{peng15arxiv}.

As all feature representations also HOG seek a reduction of information in order to arrive at a more compact representation of the visual input that is more robust to nuisances such as noise and illumination. It is specified as a mapping of an image into the HOG space. The resulting representation is then typically further used in classification or matching approaches to solve computer vision tasks.

While HOG is only defined as a feed-forward computation and introduces an information bottleneck, sometimes we desire to invert this pipeline for further analysis. E.g. previous work has tried visualize HOG features by solving an pre-image problem \cite{vondrick2013hoggles,kato2014image}. Given a HOG representation of an unobserved input image, the approaches try to estimate an image that produces the same HOG representation and is close to the original image. This has been addressed by sampling approach and approximation of the HOG computation in order to circumvent the problem of the non-invertible HOG computation. Another example, is pose estimation based on 3D models \cite{xiang_wacv14,Aubry14,2015arXiv150305038P,stark10bmvc} that exploits renderings of 3D models in order to learn a pose prediction model. Here the HOG computation is followed up by a Deformable Part Model \cite{felzenszwalb2010object} or simplified versions that related to the Exemplar Model \cite{malisiewicz-iccv11}. Typically, these methods employ sampling based approaches in order to render discrete view-points that are then used in a learning-based scheme to match to images.

In our work, we investigate directly computing the gradient of the HOG representation which then can be used for end-to-end optimization of the input w.r.t. the desired output.
For the visualization via pre-image estimation, we observe the HOG representation and compute the gradient w.r.t. the raw pixels of the input image. For pose estimation we consider the whole pose scoring pipeline of \cite{Aubry14} that constitutes a model with multiple parts and a classifier on top of the HOG representation. Here we show how to directly maximize the pose scoring function by computing the gradient w.r.t. the pose parameters. In contrast to the previous approach, we do not reply on pre-rendering views exhaustively and our pose estimation error is therefore not limited by the initial sampling.

We compare to previous works on HOG visualizations and HOG-based pose estimation using rendered images. By using our approach of end-to-end optimization via differentiation of the HOG computation, we improve over the state of the art on both tasks.

\section{Related Work}

The HOG feature representation is widely used in many computer vision based applications. Despite its popularity, its appearance in the objective functions usually makes the optimization problem hard to operate where it is treated as a non-differentiable function \cite{huang2011local, xiong2013supervised}. How to invert the the feature descriptor to inspect its original observation invokes a line of works of feature inversion and feature visualization (pre-image) problem. There are plenty of works on this interesting topic. Given the HOG features of a test image, Vondrick et al. \cite{vondrick2013hoggles} tried in their baseline to optimize the objective with HOG involved by the numerical derivatives but failed to get reasonable results, thus in their proposed method the inversion is done by learning a paired dictionary of features and the corresponding images. Weinzaepfel et al. \cite{weinzaepfel2011reconstructing} attempted to reconstruct the pre-image of the given SIFT descriptors \cite{lowe1999object} based on nearest neighbor search in a huge database of images for patches with the closet descriptors. Kato et al. \cite{kato2014image} study the problem of pre-image estimation of the bag-of-words features and they rely on a large-scale database to optimize the spatial arrangement of visual words. Although these and other related works provide different ways to approximately illustrate the characteristic of the image features, we nearly have not seen the work directly addressing the differentiable form of the feature extraction procedure. In contrast, our approach contributes to make the differentiation of HOG descriptor practical such that it can be easily plugged into the computer vision pipeline to enable direct end-to-end optimization and extension to hybrid MCMC schemes \cite{kulkarnipicture, kulkarni2015deep}. One most relevant work to ours is from Mahendran et al. \cite{mahendran15understanding}, which inverts feature descriptors (HOG \cite{felzenszwalb2010object}, SIFT \cite{lowe1999object}, and CNNs \cite{krizhevsky2012imagenet}) for a direct analysis of the visual information contained in representations, where HOG and SIFT are implemented by Convolutional Neural Networks (CNNs). However, their approach contains an approximation to the orientation binning stage of HOG/SIFT and includes two strong natural image priors in the objective function with some parameters need to be estimated from training set. Instead in our work, we do not have any approximation in the HOG pipeline and no training is needed.

Despite deep-learning based features are in fashion these years, there are plenty of applications using HOG, in particular the Examplar LDA \cite{hariharan2012discriminative} for the pose estimation task with rendered/CAD data \cite{Aubry14, lim2013parsing, christopher2015enrich}. In \cite{dong2015domain}, slightly-modified SIFT (gradient-histogram-based as HOG) can beat CNNs in feature matching task. In this paper, we specifically demonstrate the application of our $\nabla$HOG on the pose estimation problem for aligning 3D CAD models to the objects on 2D real images, we briefly review some recent research works here. Lim et al. \cite{lim2013parsing} assume the accurate 3D CAD model of the target object is given, based on the discretized space of poses for initialization they estimate the poses from the correspondences of LDA patches between the real image and the rendered image of CAD model. Aubry et al. \cite{Aubry14} create a large dataset of CAD models of chair objects, with rendering each CAD model from a large set of viewpoints they train the classifiers of discriminative exemplar patches in order to find the alignment between the chair object on the 2D image and the most similar CAD model of the certain rendering pose. In additional to the discrete pose estimation scheme as \cite{Aubry14}, there has been works on continuous pose estimation \cite{2011SongVisua, christopher2015enrich, 2015arXiv150305038P}. For instance, Pepik et al. \cite{2015arXiv150305038P} train a continuous viewpoint regressor and also the RCNN-based \cite{girshick14CVPR} key-point detectors which are used to localize the key-points on 2D images in an object class specific fashion, with the correspondences between the key-points on the 2D image and 3D CAD model, they estimate the pose of the target object. However, for these current state-of-the-art approaches most of them need to collect plenty of data to train the discriminative visual element detectors or key-point detectors for the matching, or to render many images of CAD models of various viewpoints in advance. Instead, our proposed method manages to combine the $\nabla$HOG based exemplar LDA model with the approximate differentiable renderer from \cite{loper2014opendr} which enable us to have directly end-to-end optimization for the pose parameters of the CAD model in alignment with the target object on the real images.

\section{$\nabla$HOG} \label{sec:method}

\begin{figure*}
\begin{center}
\includegraphics[width=0.675\paperwidth]{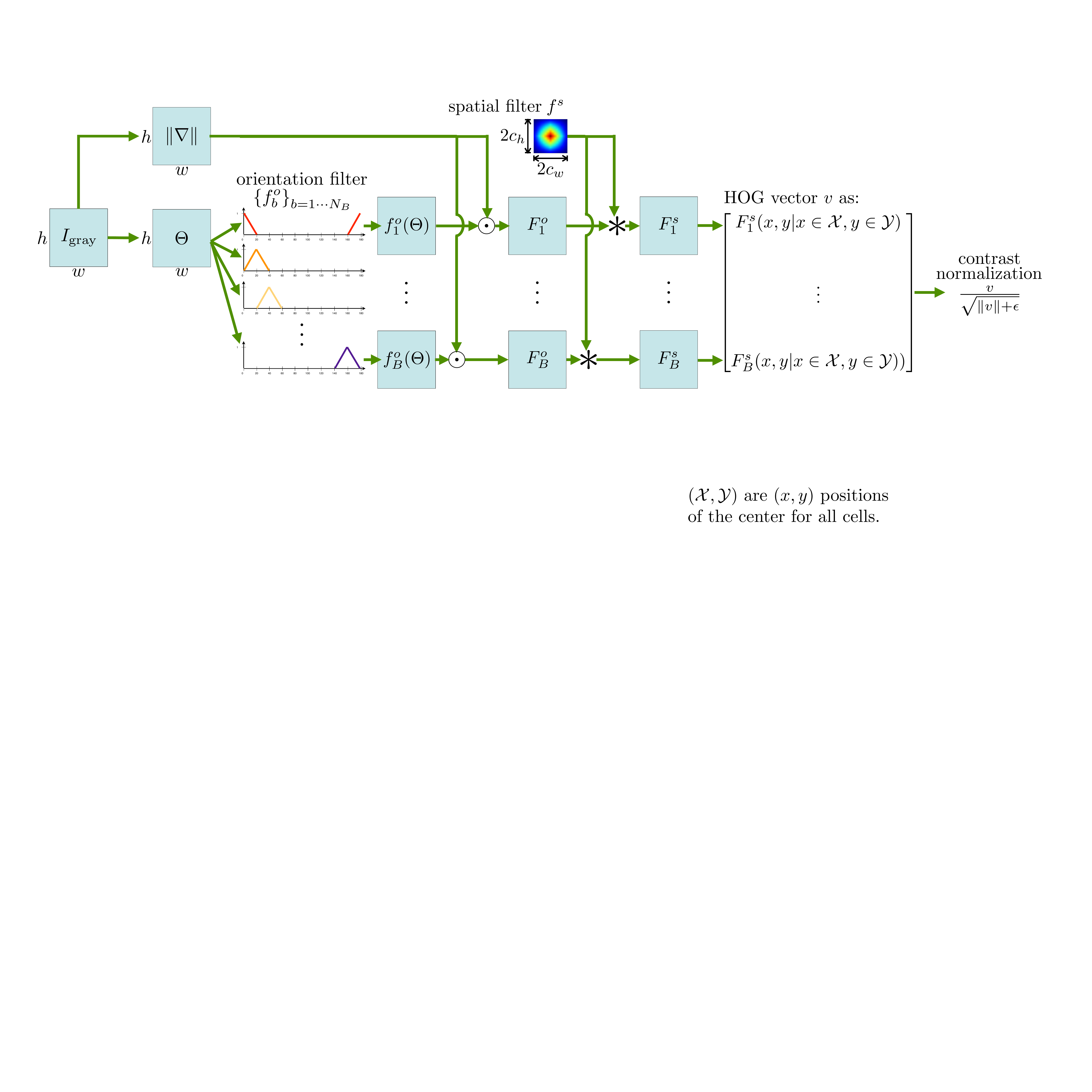}
\end{center}
\caption{Visualization of the implementation procedure for our $\nabla$HOG method. }
\label{fig:hog_visualize}
\end{figure*}

Here we describe how we achieve the derivative of the HOG descriptor. In the original HOG computation, there are few sequential key-components, including 1) computing gradients, 2) weighted vote into spatial and orientation cells, and 3) contrast normalization over overlapping spatial blocks. In our implementation we follow the same procedure. For each part we argue for piecewise differentiability. The differentiability of the whole pipeline follows from the chain rule of differentiation. Figure~\ref{fig:hog_visualize} shows an overview of the computations involved in the HOG feature computation pipeline which we describe in details in the following.

\subsection{Gradients Computation}
If a color image $I \in \mathbb{R}^{w\times h \times 3}$ is given, we first compute its gray-level image:
\begin{equation}
\label{eq:gray}
I_{\text{gray}} = I(:,:,0)\ast 0.299+I(:,:,1)\ast 0.587+I(:,:,2)\ast 0.114
\end{equation}
Then we follow the best setting for gradient computation as in Dalal's approach \cite{dalal2005histograms}, to apply the discrete derivative $1$--D $\left[ -1, 0, 1\right]$ masks on both horizontal and vertical directions without Gaussian smoothing. We denote the gradient maps on horizontal and vertical directions as $G_x$ and $G_y$, while the magnitude $\left \| \nabla  \right \|$ and direction $\Theta$ of gradients can be computed by:
\begin{align}
\label{eq:grad}
\begin{split}
\left \| \nabla  \right \| &= \sqrt{G_x^2+G_y^2}\\
\Theta &=\arctan(G_y, G_x)
\end{split}
\end{align}
Note that here we use unsigned orientations such that the numerical range of the elements in $\left \| \nabla  \right \| \in \left [ 0, 180 \right ]$. The $L2$ norm is denoted by $\left \| \cdot \right \|$ through this paper for consistency.
\paragraph{Differentiability:}
The conversion to gray as well as the derivative computation via linear filtering are linear operations and therefore differentiable.
$\arctan$ is differentiable in $\mathbb{R}$ and the gradient magnitude $\left \| \nabla  \right \|$ is differentiable due to the chaining of the differentiable squaring function and the square root over values in $\mathbb{R}^+$.

\subsection{Weighted vote into spatial and orientation cells}
After we have the magnitude and direction of the gradients, we proceed to do the weighted vote of gradients into spatial and orientation cells which provides the fundamental nonlinearity of the HOG representation. The cells are the local spatial regions where we accumulated the local statistics of gradients by the histogram binning of their orientations. Assume we divide the image region into $N^c_w \times N^c_h$ cells of size $c_w \times c_h$, for each pixel located within the cell we compute the weighted vote of its gradient orientation to an orientation histogram (In our setting we use the same setting as Dalal's to have the histogram of $9$ bins spaced over $0^{\circ} - 180^{\circ}$ which ignores the sign of the gradients).

Normally for each cell its orientation histogram is represented in a $1$--D vector of length $B$ (number of bins), but this operation will miss the positions of the pixels which contribute to the histogram. This does not lead to a formulation that allows for derivation of the HOG representation with respect to different pixel positions. Our main observation here is to view each orientation binning as a filter $f^o_b$ applied to each location in the gradient map. We store the filtered results in $F^o \in \mathbb{R}^{w \times h \times B}$. Analogously, the pixel-wise orientational filters $\left\{ f_b^o \right\}_{b=1 \cdots B}$ are chosen to follow the bi-linear interpolation scheme of the gradients in neighboring orientational bins:
\begin{eqnarray}
\label{eq:oriF}
f_b^o (\Theta) &=& \text{clip}^{max=1}_{min=0}( 1- \left |\Theta- \mu_b  \right | \times \frac{B}{180} ) \\
F^o_b &=& \left \| \nabla  \right \| \odot  f_b^o(\Theta),\quad \forall b \in 1\cdots B
\end{eqnarray}
where $\mu_b$ is the central value of orientation degree for filter $f_b^o$, $\text{clip}^{max=1}_{min=0}$ function clamps the numerical range  between $1$ and $0$, and $\odot$ is an element-wise multiplication. (Note that for the first filter $f_1^o$ we  also take care of the numerical range. See the visualization shown in Figure~\ref{fig:hog_visualize}.)

We have the $F^o$ for orientational binning, we then apply spatial binning for each cell. Here as in the Dalal's method, to reduce the aliasing, for each pixel location it will contribute to its $4$ neighboring cells proportional to the distances to the centers of those cells, in another word, the votes are interpolated bilinearly. Following the similar trick as in orientational binning, by creating a $2c_w \times 2c_h$ bilinear filter $f^s$ where its maximum value $1$ is in the center with decreasing values toward four corners to minimum value $0$, as shown in Figure~\ref{fig:hog_visualize}, we convolve it with all $F^o_b$ to get the spatial filtered results $F^s_b$:

\begin{equation}
\label{eq:spaF}
F^s_b = F^o_b \ast f^s,\quad \forall b \in 1\cdots B
\end{equation}
then the spatial binning for cells can be easily fetched from:
\begin{equation}
\label{eq:spaFsample}
F^s_b(x,y|x\in\mathcal{X}, y\in\mathcal{Y}),\quad \forall b \in 1\cdots B
\end{equation}
where $(\mathcal{X},\mathcal{Y})$ are the $(x,y)$ coordinates of the centers for all cells.

Note that till here when you concatenate $v = \left\{ F^s_b(x,y|x\in\mathcal{X}, y\in\mathcal{Y}) \right\}_{b=1\cdots B}$ then actually we get exactly the same representation as from original HOG approach.

\paragraph{Differentiability}
By re-representing the data, we have shown that the histograming and voting procedure of the HOG approach can be viewed as linear filtering operations followed up by a summation. Both steps are differentiable.

\subsection{Contrast normalization}
In the original procedure of Dalal's HOG descriptor,  contrast normalization is performed on every local region of size $3\times 3$ cells, which they call \textit{blocks}. As many recent applications that we are interested in \cite{Aubry14,Aubry13,kato2014image,vondrick2013hoggles,felzenszwalb2010object} do not use blocks, we do not consider them either in our  implementation. 
While this step is possible to incorporate, it would also lead to increased computational costs due to multiple representation of the same cell. We instead only use the global normalization by using the robust $L2\text{-}norm$. Given the HOG representation $v$ from previous steps, the global contrast normalization can be written as:
\begin{equation}
\label{eq:norm}
v_{\text{normalized}} = \frac{v}{\sqrt{\left \| v \right \|+ \epsilon}}
\end{equation}
where $\epsilon$ is a small positive constant. 

\paragraph{Differentiability:}
This is a chain of differentiable functions and therefore the whole expression is differentiable.

\paragraph{Difference to Original HOG}
While there is a large diversity of HOG implementations available by now, we summarize the two main difference to the standard one as proposed in \cite{dalal2005histograms}:
First, the original HOG compute the the gradients on different color channels and apply the maximum operator on the magnitudes over all channels to get the gradient map. In our implementation we simply first transform the color image into gray scale and compute the gradient map directly. 
Second, we do not include the local contrast normalization for every overlapping spatial blocks. But we do include the global, robust $L2$ normalization.

\subsection{Implementation}

In the above equations (Eq.~\ref{eq:gray}, \ref{eq:grad}, \ref{eq:oriF}, \ref{eq:spaF}, \ref{eq:norm}) all the operations are (piecewise-) differentiable (summation, multiplication, division, square, square root, arc-tangent, clip), with the use of the chain rule, our overall HOG implementation is differentiable on each pixel position. 
Overall, this is not surprising as visual feature representations are designed to vary smoothly w.r.t. to small changes in the image.
We have implemented this version of the HOG descriptor by using the Python-based auto-differentiation package \textit{Chumpy} \cite{Chumpy}, which evaluates an expression and its derivatives with respect to its inputs. The package and our extension integrate with the recently proposed Approximate Differentiable Renderer OpenDR \cite{loper2014opendr}. We will make our implementation publicly available in the near future.
\section{Experimental Results}

\subsection{Reconstruction from HOG descriptors}
We evaluate our proposed $\nabla$HOG method on the image reconstruction task based on the feature descriptors. We are interested in this task since it provides a way to visualize the information carried by the feature descriptors and open the opportunity to examine the feature descriptor itself instead of based on the performance measures of certain tasks as proxies. There is already prior work on this problem. \cite{kato2014image, vondrick2013hoggles, d2012beyond} focus on different feature representations such as Bag-of-Visual-Words (BoVW), Histogram of Orientated Gradients (HOG), and Local Binary Descriptors (LBDs). However, state-of-the-art approaches typically need to use large-scale image bases for learning the reconstruction.

\paragraph{Objective}
As we have derived the gradient of the HOG feature w.r.t. the input, we can -- given a feature vector -- directly optimize for the reconstruction of original image without any additional data needed. To define the problem more formally, let $I \in \mathbb{R}^{X\times Y}$ be an image and its HOG representation as $\phi(I)$, we optimize to find the reconstructed image $\hat{I}$ whose HOG features $\phi(\hat{I})$ have the minimum euclidean distance $E$ to $\phi(I)$:
\begin{align}
\begin{split}
\hat{I} &= \argmin_{\hat{I}\in \mathbb{R}^{X \times Y}} E \\
&= \argmin_{\hat{I}\in \mathbb{R}^{X \times Y}} \left \|  \phi(I) - \phi(\hat{I}) \right \|
\end{split}
\label{eq:reconstruct}
\end{align}
The option to approach the problem in this way was mentioned in \cite{vondrick2013hoggles}, however there was no result achieved as numerical differentiation is very computational expensive in this setting. Direct optimization is facilitated for the first time using our $\nabla$HOG approach.

An overview of our approach is shown in Figure \ref{fig:teaser}.
We compute derivatives $\frac{\partial E}{\partial i_{x,y}}$ with respect to the intensity values $i_{x,y}$ of all the pixel positions $(x,y)$ on $\hat{I}$ via auto-differentiation. By gradient-based optimization we are able to find a local minimum of $E$ and corresponding reconstructed image $\hat{I}$. In order to regularize our estimation, we introduce a smoothness prior that penalizes gray value changes of adjacent pixels. Intuitively, this encourages propagation of information into areas without strong edges for which no signal from the HOG features is available.
\begin{equation}
\hat{I} = \argmin_{\hat{I}\in \mathbb{R}^{X \times Y}} \left \|  \phi(I) - \phi(\hat{I}) \right \| + \xi \sum_{p, q \in \mathcal{N}} \left \| i_p - i_q \right \|  
\label{eq:reconstruct_add_smooth}
\end{equation}
where $p, q \in \mathcal{N}$ means that pixel $p$ and $q$ are neighbors, and $\xi$ is the weight for the smoothness term which we usually set to a big number as \SI{e+2}{} in our experiments. Although we give a high weight for the smoothness term, it will only play a key role in the first few iterations of the optimization procedure then the euclidean distance $E$ will dominate to find the local minimum.  

The evaluation is based on the image reconstruction dataset proposed in \cite{kato2014image} which contains $101$ images for all the categories from Caltech $101$ dataset \cite{fei2007learning} and all have a resolution of $128 \times 128$. 
We compare our method with few state-of-the-art baselines on image reconstruction from feature descriptions: the \textbf{BoVW} method from \cite{kato2014image}, the \textbf{HOGgles} method from \cite{vondrick2013hoggles}, also \textbf{CNN-HOG} and \textbf{CNN-HOGb}(CNN-HOG with bilinear orientation assignments) from \cite{mahendran15understanding}.

Note that our $\nabla$HOG described in Section~\ref{sec:method} is based on \textbf{Dalal's}-type HOG\cite{dalal2005histograms}, while for HOGgles/CNN-HOG/CNN-HOGb baselines they are using \textbf{UoCTTI}-type HOG\cite{felzenszwalb2010object} which additionally considers directed gradients. To have a fair comparison, we also implement UoCTTI HOG under our proposed framework.

\begin{figure}
\begin{center}
\subfigure[$\nabla$HOG multi-scale]{\label{fig:multi-scale}\includegraphics[height=0.62\linewidth]{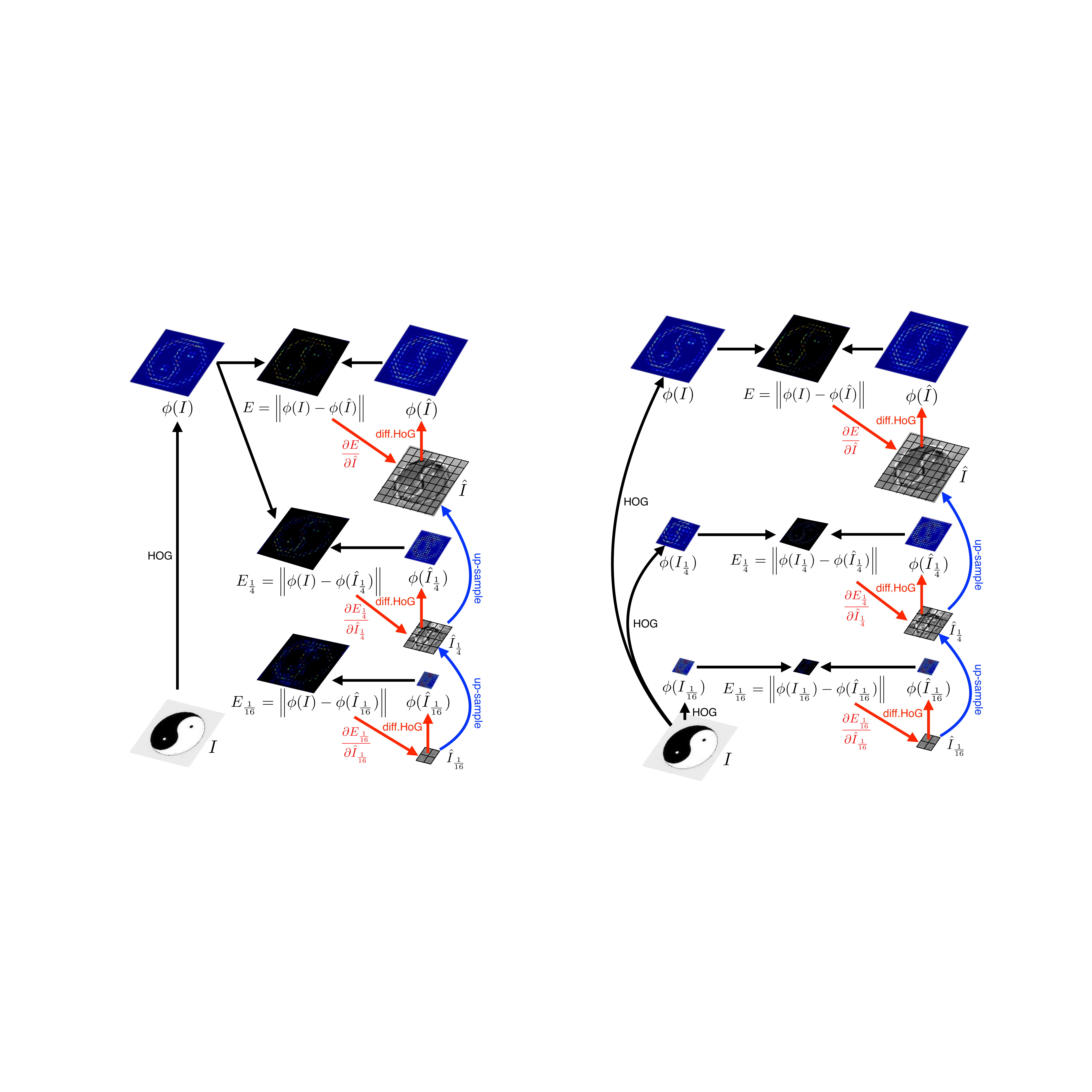}}
\subfigure[$\nabla$HOG multi-scale-more]{\label{fig:multi-scale-more}\includegraphics[height=0.62\linewidth]{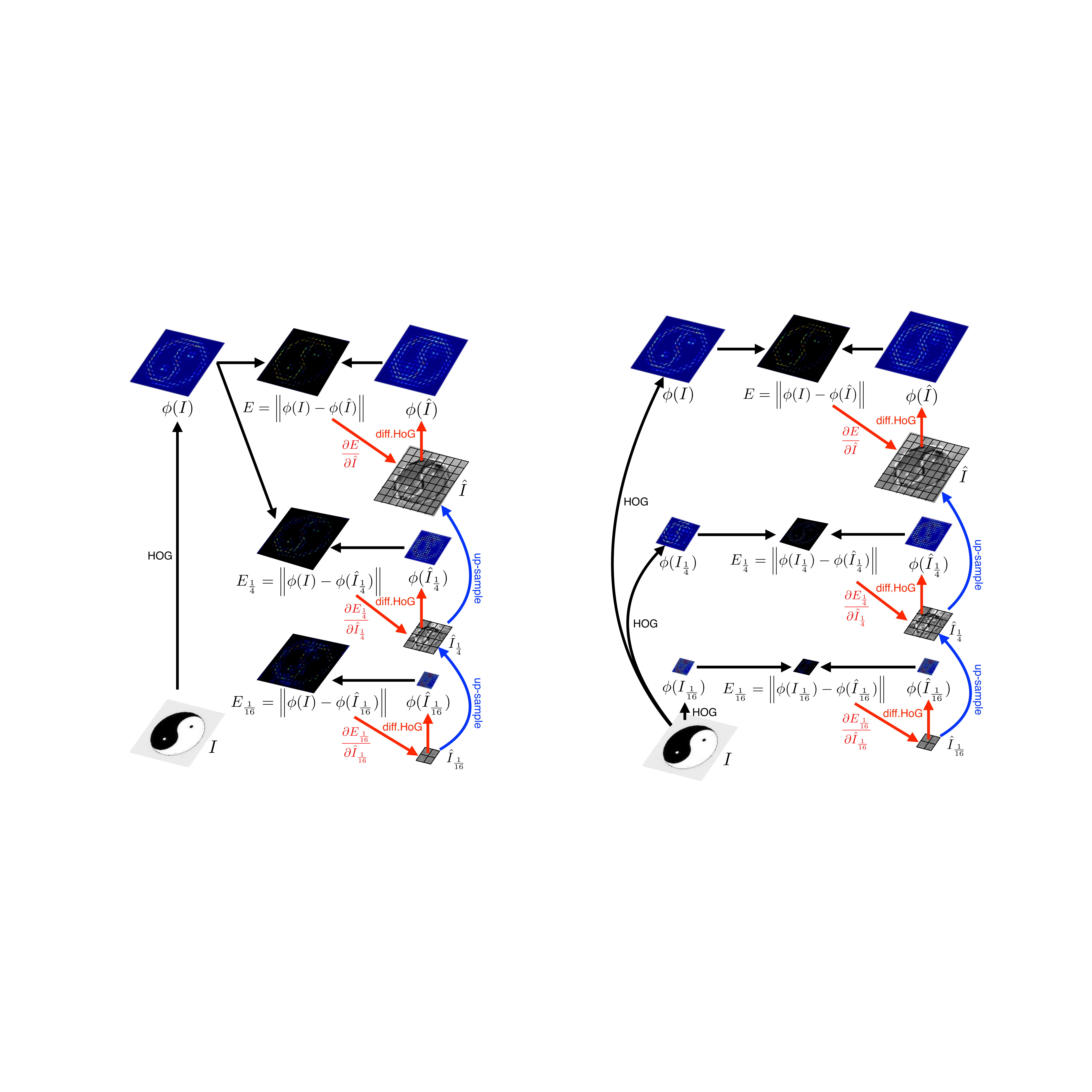}}
\end{center}
\caption{Visualizations of variants of the proposed method for the task of image reconstruction from feature descriptors.}
\label{fig:reconstruct_vis}
\end{figure}

We propose two additional variants for reconstruction that exploit multi-scale information as shown in Figure~\ref{fig:reconstruct_vis}.

\paragraph{$\nabla$HOG multi-scale} We use the single scale HOG descriptor as input, but we first reconstruct $\hat{I}_{\frac{1}{s}}$ with $s$ times smaller resolution than $I$ (the cell size for $\phi(\hat{I}_{\frac{1}{s}})$ is $\frac{1}{\sqrt{s}}$ of the original one used for $\phi(I)$, $s \in \{ 4, 16, 64\}$ in our experimental setting.). After few iterations of updates in optimization process, we up-sample $\hat{I}_{\frac{1}{s}}$ to higher resolution and continue the reconstruction procedure. These steps are repeated until we reach the initial resolution of $I$.  

\paragraph{$\nabla$HOG multi-scale-more} We use the multi scale HOG vectors of the original image $I$ as the input. For the reconstruction on different scale $s$, the corresponding HOG descriptor $\phi(I_{\frac{1}{s}})$ extracted on the same scale is used in the euclidean distance $E$, as shown in Figure~\ref{fig:multi-scale-more}. As additional HOG descriptors are computed from the original image at different scales, we use more information than in the original setup and therefore the results of the ``multi-scale-more'' approach cannot be directly compared to prior works. 

The optimization is done based on the nonlinear optimization using Powell's dogleg method \cite{lourakis2005levenberg} which is implemented in \textit{Chumpy} \cite{Chumpy}.
Example results of the multi scale approaches can be seen in Table~\ref{tab:multi-scale-figures}.

\begin{table}[ht] 
\centering
\begin{tabular}{p{0.1cm} p{1cm} p{1cm} p{1cm} p{1cm} p{1.5cm}}
& test & $1/64$ & $1/16$ & $1/4$ & $1/1$ \\
\raisebox{+2\totalheight}{(a)} &
\includegraphics[width=0.165\columnwidth]{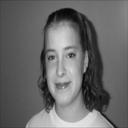} & 
\includegraphics[width=0.165\columnwidth]{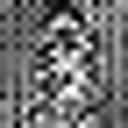} &
\includegraphics[width=0.165\columnwidth]{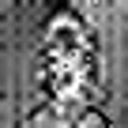} &
\includegraphics[width=0.165\columnwidth]{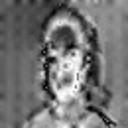} &
\includegraphics[width=0.165\columnwidth]{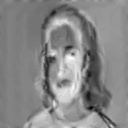} \\
\raisebox{+2\totalheight}{(b)} &
\includegraphics[width=0.165\columnwidth]{images/gray/001.jpg} & 
\includegraphics[width=0.165\columnwidth]{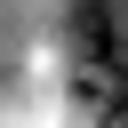} &
\includegraphics[width=0.165\columnwidth]{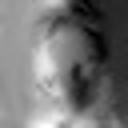} &
\includegraphics[width=0.165\columnwidth]{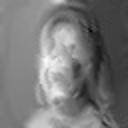} &
\includegraphics[width=0.165\columnwidth]{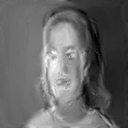} \\
& test & iter-$0$ & iter-$1$ & iter-$2$ & iter-final \\
\raisebox{+2\totalheight}{(c)} &
\includegraphics[width=0.165\columnwidth]{images/gray/001.jpg} & 
\includegraphics[width=0.165\columnwidth]{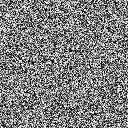} &
\includegraphics[width=0.165\columnwidth]{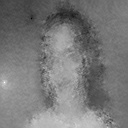} &
\includegraphics[width=0.165\columnwidth]{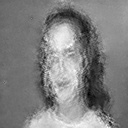} &
\includegraphics[width=0.165\columnwidth]{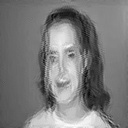} \\
\end{tabular}
\caption{Example results for (a)(b) \textbf{$\nabla$HOG multi-scale} and \textbf{$\nabla$HOG multi-scale-more} in which both are based on Dalal-HOG\cite{dalal2005histograms}; and (c) for \textbf{$\nabla$HOG} on UoCTTI-HOG\cite{felzenszwalb2010object}. }
\label{tab:multi-scale-figures}
\end{table}

\paragraph{Results}

In order to quantify the performance of image reconstruction, different metrics have been proposed in prior works. For instance, in \cite{kato2014image} the mean squared error of raw pixels is utilized, while in \cite{vondrick2013hoggles} the cross-correlation is chosen to compare the similarity between the reconstructed image and the original one. In addition to using cross-correlation as the metric for qualitative evaluation, we also investigate different choices used by the research works on the problem of image quality assessment (IQA), including mutual information and \textbf{Structural Similarity (SSIM)} \cite{wang2004image}. In particular, mutual information measures the mutual dependencies between images hence gives another metric for similarities, while SSIM measures the degradation of structural information for the distorted/reconstructed image from the original one, under the assumption that human visual perception is adapted to discriminate the structural information from the image.

\begin{table*}[ht]
\begin{tabular}{|l|l|c|c|c|}
\hline
\multicolumn{2}{|c|}{Method} & cross correlation & mutual information & structural  similarity (SSIM) \cite{wang2004image}\\ \hline 
& BoVW \cite{kato2014image}  & 0.287 & 1.182 & 0.252 \\ \hline \hline \hline

\multirow{4}{*}{\rotatebox[origin=c]{90}{\begin{tabular}{@{}c@{}}UoCTTI  \\ HOG \cite{felzenszwalb2010object}\end{tabular}}} & HOGgles \cite{vondrick2013hoggles} & 0.409 &1.497 &  0.271  \\ \cline{2-5}
& CNN-HOG \cite{mahendran15understanding} & 0.632 & 1.211 &  0.381  \\ \cline{2-5}
& CNN-HOGb \cite{mahendran15understanding}  & 0.657 & 1.597 &  0.387  \\ \cline{2-5}
& our $\nabla$HOG  (single scale)  & \textbf{0.760} & \textbf{1.908} &  \textbf{0.433}  \\ \hline \hline \hline

\multirow{9}{*}{\rotatebox[origin=c]{90}{\begin{tabular}{@{}c@{}}Dalal's  \\ HOG \cite{dalal2005histograms}\end{tabular}}} & our $\nabla$HOG  (single scale) & 0.170 & 1.464 &  0.301   \\ \cline{2-5}
& our $\nabla$HOG  (multi-scale: $\frac{1}{64}$) & 0.058 & 1.444 &  0.121  \\ \cline{2-5}
& our $\nabla$HOG  (multi-scale: $\frac{1}{16}$) & 0.076 & 1.470 &  0.147  \\ \cline{2-5}
& our $\nabla$HOG  (multi-scale: $\frac{1}{4}$) & 0.108 & 1.458 &  0.221  \\ \cline{2-5}
& our $\nabla$HOG  (multi-scale: $\frac{1}{1}$) & 0.147 & 1.478 &  0.293   \\ \cline{2-5}

& our $\nabla$HOG  (multi-scale-more: $\frac{1}{64}$) & 0.147 & 1.458 &  0.251  \\ \cline{2-5}
& our $\nabla$HOG  (multi-scale-more: $\frac{1}{16}$) & 0.191 & 1.502 &  0.291  \\ \cline{2-5}
& our $\nabla$HOG  (multi-scale-more: $\frac{1}{4}$) & 0.220 & 1.565 &  0.320   \\ \cline{2-5}
& our $\nabla$HOG  (multi-scale-more: $\frac{1}{1}$) & \textbf{0.236} & \textbf{1.582} &  \textbf{0.338}  \\ \hline
\end{tabular}
\caption{Comparison on the performance of reconstruction from feature descriptors.}
\label{table:reconstruct_results}
\end{table*}

We report the performance numbers from all the metrics in Table~\ref{table:reconstruct_results}. The proposed method using UoCTTI-type HOG outperforms the state-of-the-art baselines by a large margins for all metrics. Visually inspected, our proposed method can reconstruct many details in the images and also give accurate estimate on gray-scale values if using UoCTTI HOG. Please note again, our method does not need any additional data for training while in baselines training is necessary.

\subsection{Pose estimation}\label{sec:pose}
\begin{figure}
\begin{center}
\includegraphics[width=1\linewidth]{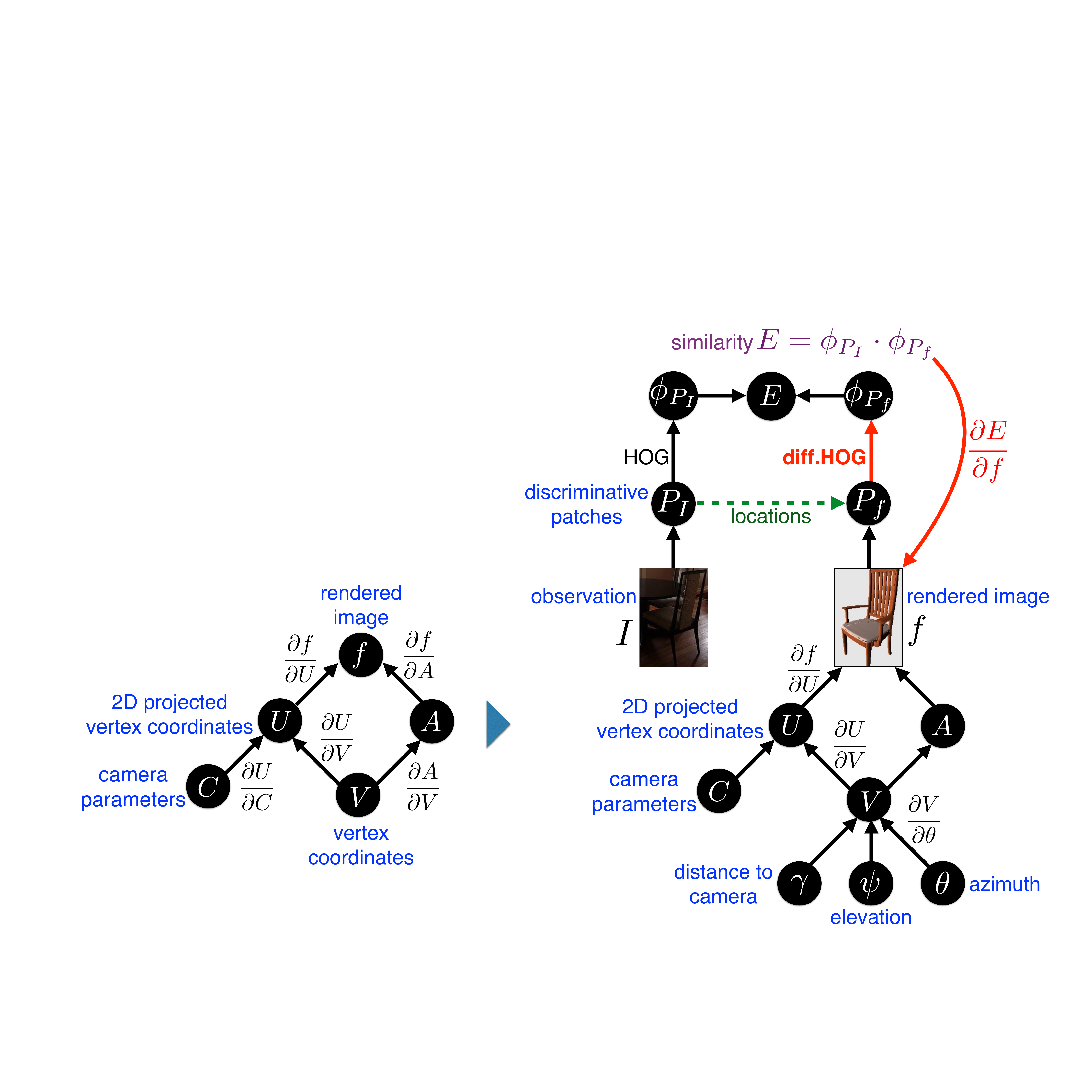}
\end{center}
\caption{(left) The differentiable rendering procedure from openDR \cite{loper2014opendr}. (right) The visualization of our model for pose estimation.}
\label{fig:pose_model}
\end{figure}

We also evaluate our $\nabla$HOG approach on a pose estimation task where  $3$D CAD models have to be aligned to objects in $2$D images. We build on openDR \cite{loper2014opendr} which is an approximate differentiable renderer. It parameterizes the forward graphics model $f$ based on vertices locations $V$, per-vertex brightness $A$ and camera parameters $C$, which is shown on the left part of Figure~\ref{fig:pose_model}, where $U$ is for the $2D$ projected vertex coordinate position. Based on the auto-differentiation techniques, openDR provides a way to derive the derivatives of the rendered image observation with respect to the parameters in the rendering pipeline.

\paragraph{Approach}
We extend openDR in the following ways as illustrated in Figure~\ref{fig:pose_model}: \textbf{1)} We parameterize the vertices locations $V$ of CAD models by three parameters: azimuth $\theta$, elevation $\psi$, and distance to the camera $\gamma$; \textbf{2)} During the pose estimation procedure, as in \cite{Aubry14}, the matching between the objects on real images and the rendered images from the CAD models are addressed by the similarities between the HOG descriptors of the visual discriminative elements extracted from them. The detailed procedure of extracting visual discriminative elements is discussed in \cite{Aubry14}. In our method, we use our $\nabla$HOG method $\phi(P_f)$ for the image patches $P_f$ which have the same regions as the visual elements $P_I$ extracted from the test image $I$, and the similarity between the $P_f$ and $P_I$ is the dot product between HOG descriptors $\phi(P_I)$ of $P_I$ and $\phi(P_f)$. As shown in the right part of Figure~\ref{fig:pose_model} this similarity can be traversed back to the pose parameters $\{ \theta, \psi, \gamma \}$ and the derivatives of the similarity with respect to the pose parameters can be again  computed by the auto-differentiation, our method can directly optimize to maximize the similarity to estimate the poses.  

\begin{figure}
\begin{center}
\includegraphics[width=1\columnwidth]{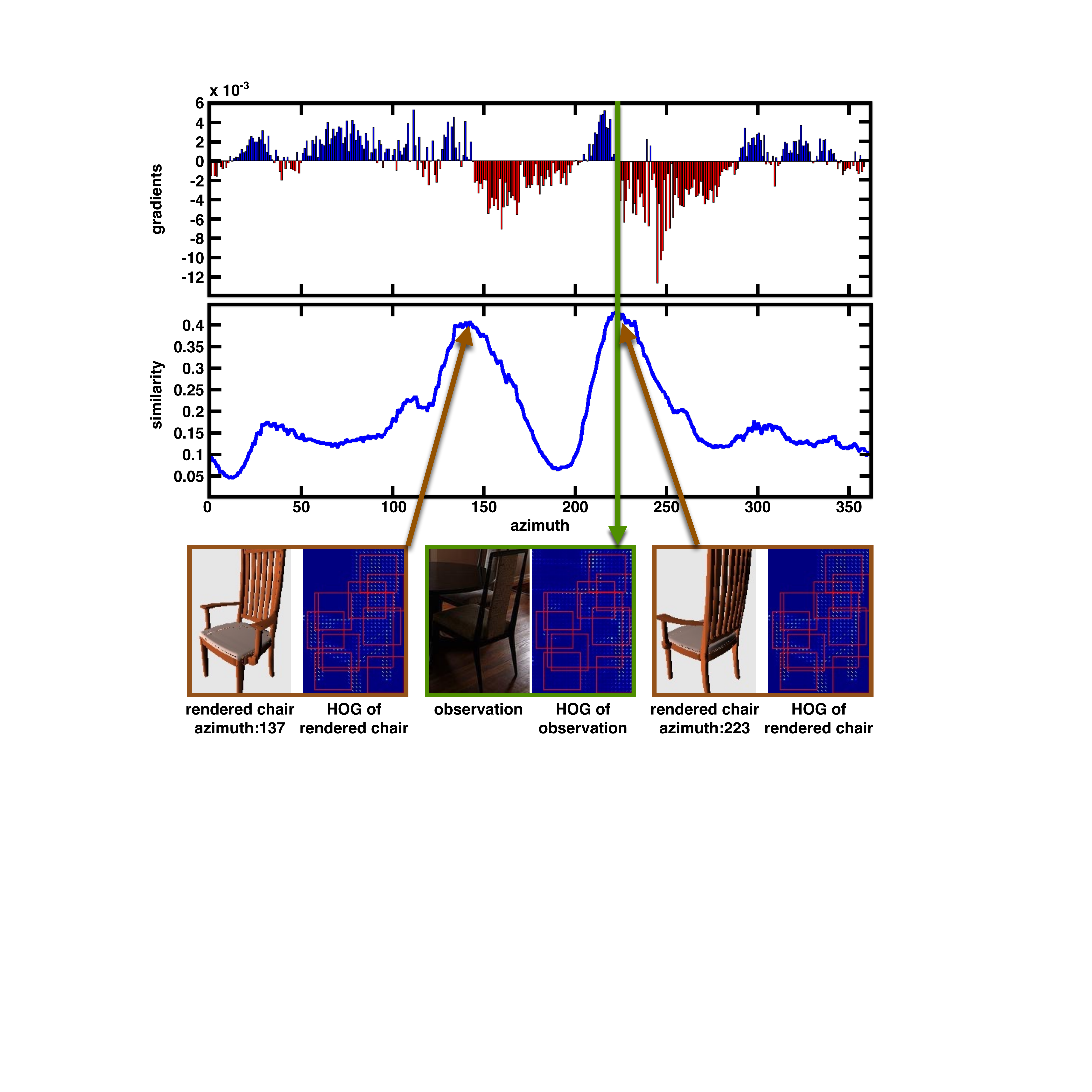}
\end{center}
\caption{Visualization of the similarity and its gradients w.r.t azimuth. The red boxes on the HOG representations are the visual discriminative patches.}
\label{fig:gradient_LDA}
\end{figure}

\paragraph{Setup}
We follow the same experimental setting as \cite{Aubry14}, where we test on the images annotated with no-occlusion, no-truncation and not-difficult of the chairs validation set on PASCAL VOC 2012 dataset \cite{Everingham15}, therefore in total $247$ chairs from $179$ images are used for the evaluation. To purely focus on evaluation of the pose estimation, we extract the object images based on their bounding boxes annotation, and resize them to have at least length of $100$ pixels on the shortest side of image size. 

The baseline \cite{Aubry14} is applied on the chair images to search over a chair CAD database of $1393$ models which includes the rendered images from $62$ different poses relative to camera for each of them, then to detect the chairs, match the styles of the chairs, and simultaneously recover their poses based on rendered images. 
We select the most confident detection for each chair together with the estimated pose.

We apply our proposed method on pose estimation by using the elevation and azimuth estimates of \cite{Aubry14} as a initialization of pose, and add few more initializations for azimuth ($8$ equidistantly distribute over $360^{\circ}$).
We use gradient descent method with momentum term for optimization in order to optimize for the azimuth parameter and interleave iterations in which we additionally optimize for the distance to camera.
In Figure~\ref{fig:gradient_LDA} we visualize an example of the similarity between the chair object on the real image and the CAD model on the rendered image, as well as  its gradients w.r.t azimuth $\theta$ (full $360^\circ$). We can see how gradients change related to different local maximums and the corresponding poses of the CAD model. 

\paragraph{Results}
In order to quantify our performance on pose estimation task, we use the continuous $3$D pose annotations from PASCAL3D+ dataset \cite{xiang_wacv14}. Following the same evaluation scheme, the view-point estimation is considered to be correct if its estimated viewpoint label is within the same interval of the discrete viewpoint space as the ground-truth annotation, or its difference with ground-truth in continuous viewpoint space is lower than a threshold. We evaluate the performance based on various settings of the intervals and thresholds in viewpoint space: $\{ 4~\text{views}/90^{\circ}, 8~\text{views}/45^{\circ}, 16~\text{views}/22.5^{\circ}, 24~\text{views}/15^{\circ} \}$. In Table~\ref{tab:pose_results} we report the performance numbers for Aubry's baseline and our proposed approach.
We are outperforming the previous best performance up to $10\%$ points on the coarse and fine measures.
Some example results which show improvements of the baseline method are shown in Table~\ref{tab:pose_figures}.

\paragraph{Discussion}
One advantage of our proposed method is that we are able to parameterize the vertexes coordinates of the CAD models by the same pose parameters as used in \cite{Aubry14}, then the differentiable rendering procedure provided by openDR \cite{loper2014opendr} and our $\nabla$HOG representations enable us to directly compute the derivatives of the similarity with respect to the pose parameters, and optimize for continuous pose parameters. In another word, for the proposed approach we do not need to discretize the parameters as \cite{Aubry14} and do not need to render images from many poses in advance for the alignment procedure either. 

\begin{table}[h]
\begin{tabular}{|c|c|c|c|c|}
\hline
                & 4 views & 8 views & 16 views & 24 views \\ \hline
Aubry et al. \cite{Aubry14}  & 47.33   & 35.39   & 20.16    & 15.23    \\ \hline
our method & 58.85   & 40.74   & 22.22    & 16.87    \\ \hline
\end{tabular}
\caption{Pose estimation results based on the groundtruth annotation from PASCAL3D+ \cite{xiang_wacv14}.}
\label{tab:pose_results}
\end{table}

\begin{table*}[ht] 
\centering
\begin{tabular}{cccccccccc}
\hline 
\hline  \raisebox{+2\totalheight}{test images} &
\includegraphics[height=0.1\linewidth]{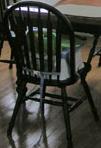} & 
\includegraphics[height=0.1\linewidth]{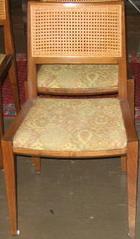} &
\includegraphics[height=0.1\linewidth]{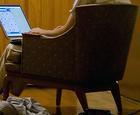} &
\includegraphics[height=0.1\linewidth]{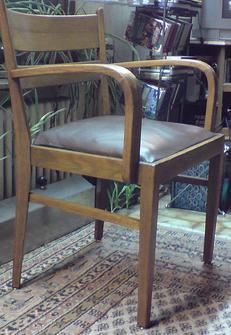} &
\includegraphics[height=0.1\linewidth]{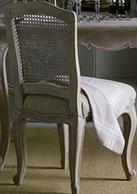} &
\includegraphics[height=0.1\linewidth]{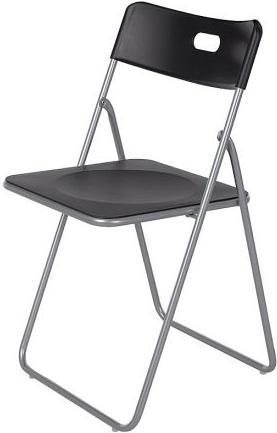} &
\includegraphics[height=0.1\linewidth]{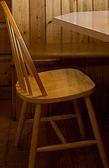} &
\includegraphics[height=0.1\linewidth]{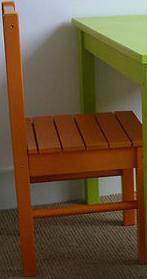} &
\includegraphics[height=0.1\linewidth]{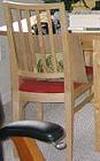}
\\ \hline 
\hline \raisebox{+2\totalheight}{Aubry et al. \cite{Aubry14}} &
\includegraphics[height=0.1\linewidth]{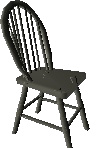} &
\includegraphics[height=0.1\linewidth]{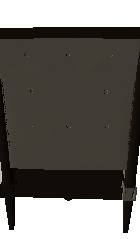} &
\includegraphics[height=0.1\linewidth]{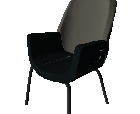} &
\includegraphics[height=0.1\linewidth]{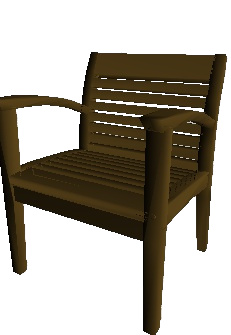} &
\includegraphics[height=0.1\linewidth]{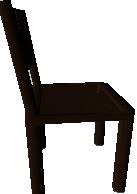} &
\includegraphics[height=0.1\linewidth]{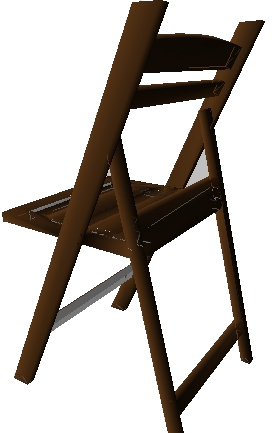} &
\includegraphics[height=0.1\linewidth]{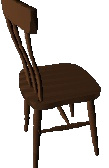} &
\includegraphics[height=0.1\linewidth]{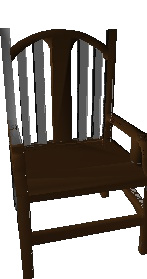} &
\includegraphics[height=0.1\linewidth]{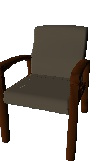}
\\ \hline 
\hline \raisebox{+2\totalheight}{our method} &
\includegraphics[height=0.1\linewidth]{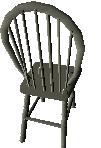} &
\includegraphics[height=0.1\linewidth]{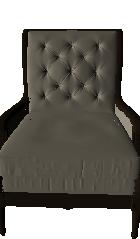} &
\includegraphics[height=0.1\linewidth]{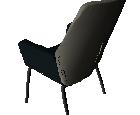} &
\includegraphics[height=0.1\linewidth]{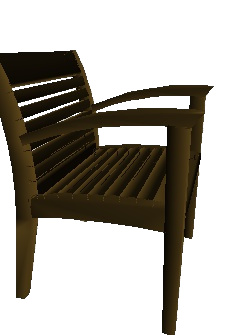} &
\includegraphics[height=0.1\linewidth]{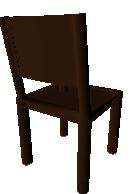} &
\includegraphics[height=0.1\linewidth]{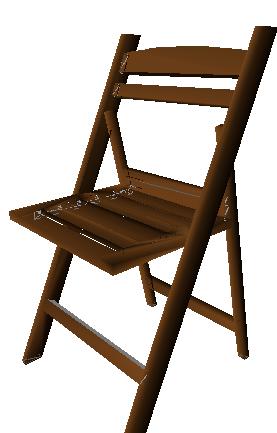} &
\includegraphics[height=0.1\linewidth]{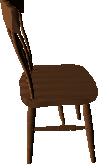} &
\includegraphics[height=0.1\linewidth]{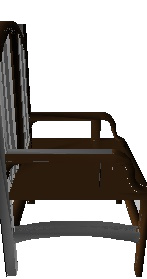} &
\includegraphics[height=0.1\linewidth]{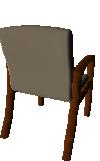}
\\ \hline \hline 
\end{tabular}
\caption{Example results for pose estimation.}
\label{tab:pose_figures}
\end{table*}

\begin{table*}[ht] 
  \resizebox{0.9\paperwidth}{!}{%
  \begin{tabular} {ll | llll | llll}
\begin{tabular}{@{}c@{}}\footnotesize{Example}\end{tabular} & \begin{tabular}{@{}c@{}}\footnotesize{HOG}\end{tabular} &
\begin{tabular}{@{}c@{}}\footnotesize{BOVW}  \\ \footnotesize{\cite{kato2014image}}\end{tabular} &
\begin{tabular}{@{}c@{}}\footnotesize{HOGgles}  \\ \footnotesize{\cite{vondrick2013hoggles}} \\ \footnotesize{UoCTTI-HOG}\end{tabular} &
\begin{tabular}{@{}c@{}}\footnotesize{CNN-HOG} \\ \footnotesize{\cite{mahendran15understanding}} \\ \footnotesize{UoCTTI-HOG}\end{tabular} &
\begin{tabular}{@{}c@{}}\footnotesize{CNN-HOGb}  \\ \footnotesize{\cite{mahendran15understanding}} \\ \footnotesize{UoCTTI-HOG}\end{tabular} &
\begin{tabular}{@{}c@{}}\footnotesize{Our $\nabla$HOG} \\ \footnotesize{(single-scale)} \\ \footnotesize{UoCTTI-HOG} \end{tabular}&
\begin{tabular}{@{}c@{}}\footnotesize{Our $\nabla$HOG}  \\ \footnotesize{(single-scale)} \\ \footnotesize{Dalal-HOG} \end{tabular} & 
\begin{tabular}{@{}c@{}}\footnotesize{Our $\nabla$HOG}  \\ \footnotesize{(multi-scale)} \\ \footnotesize{Dalal-HOG} \end{tabular} & 
\begin{tabular}{@{}c@{}}\footnotesize{Our $\nabla$HOG}  \\ \footnotesize{(multi-scale-more)} \\ \footnotesize{Dalal-HOG} \end{tabular} \\ 
\parbox[c]{4em}{\includegraphics[width=0.195\columnwidth]{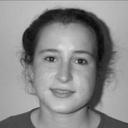}} &
\parbox[c]{4.3em}{\includegraphics[width=0.195\columnwidth]{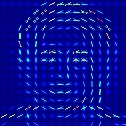}} &
\parbox[c]{4em}{\includegraphics[width=0.195\columnwidth]{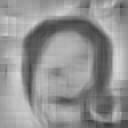}} &
\parbox[c]{4em}{\includegraphics[width=0.195\columnwidth]{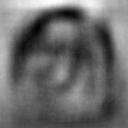}} &
\parbox[c]{4em}{\includegraphics[width=0.195\columnwidth]{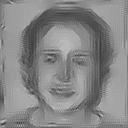}} &
\parbox[c]{4em}{\includegraphics[width=0.195\columnwidth]{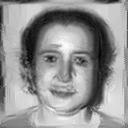}} &
\parbox[c]{4em}{\includegraphics[width=0.195\columnwidth]{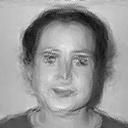}} &
\parbox[c]{4em}{\includegraphics[width=0.195\columnwidth]{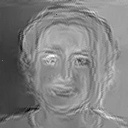}} &
\parbox[c]{4em}{\includegraphics[width=0.195\columnwidth]{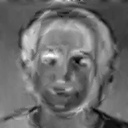}} &
\parbox[c]{4em}{\includegraphics[width=0.195\columnwidth]{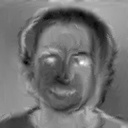}} \\ 
\parbox[c]{4em}{\includegraphics[width=0.195\columnwidth]{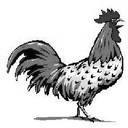}} &
\parbox[c]{4em}{\includegraphics[width=0.195\columnwidth]{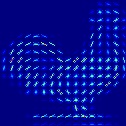}} &
\parbox[c]{4em}{\includegraphics[width=0.195\columnwidth]{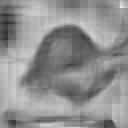}} &
\parbox[c]{4em}{\includegraphics[width=0.195\columnwidth]{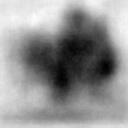}} &
\parbox[c]{4em}{\includegraphics[width=0.195\columnwidth]{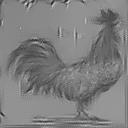}} &
\parbox[c]{4em}{\includegraphics[width=0.195\columnwidth]{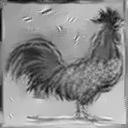}} &
\parbox[c]{4em}{\includegraphics[width=0.195\columnwidth]{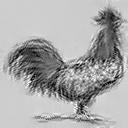}} &
\parbox[c]{4em}{\includegraphics[width=0.195\columnwidth]{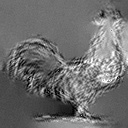}} &
\parbox[c]{4em}{\includegraphics[width=0.195\columnwidth]{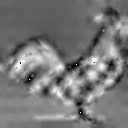}} &
\parbox[c]{4em}{\includegraphics[width=0.195\columnwidth]{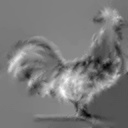}} \\
\parbox[c]{4em}{\includegraphics[width=0.195\columnwidth]{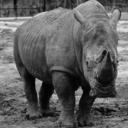}} &
\parbox[c]{4em}{\includegraphics[width=0.195\columnwidth]{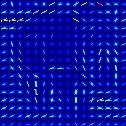}} &
\parbox[c]{4em}{\includegraphics[width=0.195\columnwidth]{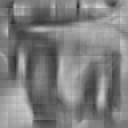}} &
\parbox[c]{4em}{\includegraphics[width=0.195\columnwidth]{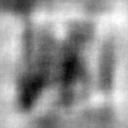}} &
\parbox[c]{4em}{\includegraphics[width=0.195\columnwidth]{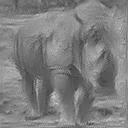}} &
\parbox[c]{4em}{\includegraphics[width=0.195\columnwidth]{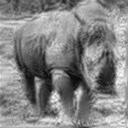}} &
\parbox[c]{4em}{\includegraphics[width=0.195\columnwidth]{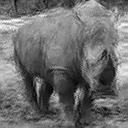}} &
\parbox[c]{4em}{\includegraphics[width=0.195\columnwidth]{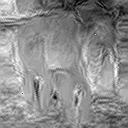}} &
\parbox[c]{4em}{\includegraphics[width=0.195\columnwidth]{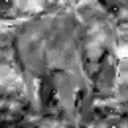}} &
\parbox[c]{4em}{\includegraphics[width=0.195\columnwidth]{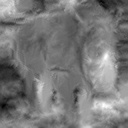}} \\
\parbox[c]{4em}{\includegraphics[width=0.195\columnwidth]{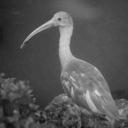}} &
\parbox[c]{4em}{\includegraphics[width=0.195\columnwidth]{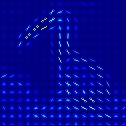}} &
\parbox[c]{4em}{\includegraphics[width=0.195\columnwidth]{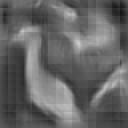}} &
\parbox[c]{4em}{\includegraphics[width=0.195\columnwidth]{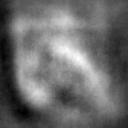}} &
\parbox[c]{4em}{\includegraphics[width=0.195\columnwidth]{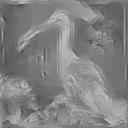}} &
\parbox[c]{4em}{\includegraphics[width=0.195\columnwidth]{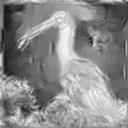}} &
\parbox[c]{4em}{\includegraphics[width=0.195\columnwidth]{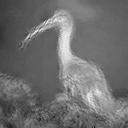}} &
\parbox[c]{4em}{\includegraphics[width=0.195\columnwidth]{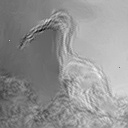}} &
\parbox[c]{4em}{\includegraphics[width=0.195\columnwidth]{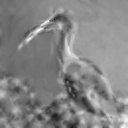}} &
\parbox[c]{4em}{\includegraphics[width=0.195\columnwidth]{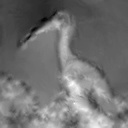}} \\
\parbox[c]{4em}{\includegraphics[width=0.195\columnwidth]{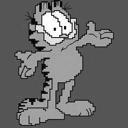}} &
\parbox[c]{4em}{\includegraphics[width=0.195\columnwidth]{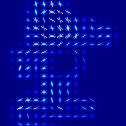}} &
\parbox[c]{4em}{\includegraphics[width=0.195\columnwidth]{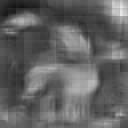}} &
\parbox[c]{4em}{\includegraphics[width=0.195\columnwidth]{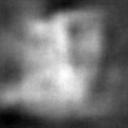}} &
\parbox[c]{4em}{\includegraphics[width=0.195\columnwidth]{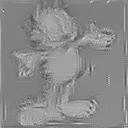}} &
\parbox[c]{4em}{\includegraphics[width=0.195\columnwidth]{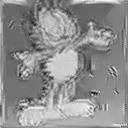}} &
\parbox[c]{4em}{\includegraphics[width=0.195\columnwidth]{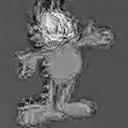}} &
\parbox[c]{4em}{\includegraphics[width=0.195\columnwidth]{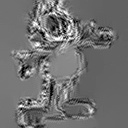}} &
\parbox[c]{4em}{\includegraphics[width=0.195\columnwidth]{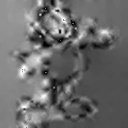}} &
\parbox[c]{4em}{\includegraphics[width=0.195\columnwidth]{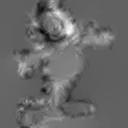}} \\
  \end{tabular}
  }
  \caption{Example results for image reconstruction from feature descriptors.}
  \label{fig:reconstruct_results}
\end{table*}

 \section{Conclusions}
We investigate the feature extraction pipeline of HOG descriptor and exploit its piecewise differentiability. Based on the implementation using auto-differentiation techniques, the derivatives of the HOG representation can be directly computed. We study two problems of image reconstruction from HOG features and HOG-based pose estimation while the direct end-to-end optimization becomes practical with our $\nabla$HOG. We demonstrate that our $\nabla$HOG-based approaches outperforms the state-of-the-art baselines for both problems.
We have demonstrated that the approach can lead to improved introspection via visualizations and improved performance via direct optimization through a whole vision pipeline.
 Our implementation is integrated into an existing auto-differentiation package as well as the recently proposed Approximately Differentiable Renderer OpenDR \cite{loper2014opendr} and is publicly available. Therefore it is easy to adopt to new tasks and is applicable to a range of end-to-end optimization problems.

\section{Acknowledgement}
We thank Matthew Loper for assistance with his great OpenDR \cite{loper2014opendr} package. We are also immensely grateful to Mathieu Aubry, Yu Xiang, Kato Hiroharu, Konstantinos Rematas, and Bojan Pepik for their help and support.

{\small
\bibliographystyle{ieee}
\bibliography{egbib}
}

\end{document}